\documentclass[runningheads]{llncs}

\usepackage[mobile]{eccv}

\usepackage{eccvabbrv}

\usepackage{graphicx}
\usepackage{booktabs}
\usepackage{wrapfig}
\usepackage{xcolor}
\usepackage[table]{xcolor}
\usepackage[font=small,labelfont=bf]{caption}
\usepackage[accsupp]{axessibility}  
\usepackage{multirow}
\usepackage{pifont}
\newcommand{\cmark}{\textcolor{green!60!black}{\ding{51}}}
\newcommand{\xmark}{\textcolor{red!70!black}{\ding{55}}}

\usepackage{caption}
\usepackage{adjustbox}
\usepackage{booktabs,colortbl}

\usepackage[breaklinks,colorlinks,citecolor=eccvblue]{hyperref}
\usepackage{hyperref}
\usepackage{amsmath,amssymb}

\begin{document}

\title{SceneAdapt: Scene-aware Adaptation \\of Human Motion Diffusion} 

\titlerunning{Abbreviated paper title}


\authorrunning{F.~Author et al.}

\title{SceneAdapt: Scene-aware Adaptation \\of Human Motion Diffusion} 

\titlerunning{SceneAdapt}

\author{Jungbin Cho\inst{1,2}\thanks{Equal contribution.} \and
Minsu Kim\inst{1}$^{\star}$ \and 
Jisoo Kim\inst{1} \and
Ce Zheng\inst{2} \and 
László A. Jeni\inst{2} \and
Ming-Hsuan Yang\inst{3,4} \and 
Youngjae Yu\inst{5} \and 
Seonjoo Kim\inst{1}}

\authorrunning{J.~Cho et al.}

\institute{Yonsei University \and
Carnegie Mellon University \and
UC Merced \and
Google DeepMind \and
Seoul National University}

\maketitle
\begin{abstract}
Human motion is inherently diverse and semantically rich, while also shaped by the surrounding scene. However, existing motion generation approaches fail to generate semantically diverse motion while simultaneously respecting geometric scene constraints, since constructing large-scale datasets with both rich text-motion coverage and precise scene interactions is extremely challenging. In this work, we introduce \textbf{SceneAdapt}, a two-stage adaptation framework that enables semantically diverse, scene-aware human motion generation from text \emph{without} large-scale paired text--scene--motion data. Our key idea is to use motion inbetweening, a learnable proxy task that requires no text, as a bridge between two disjoint resources: a text-motion dataset and a scene-motion dataset. By first adapting a text-to-motion model through inbetweening and then through scene-aware inbetweening, SceneAdapt injects geometric scene constraints into text-conditioned generation while preserving semantic diversity. To enable adaptation for inbetweening, we propose a novel \emph{Context-aware Keyframing} (CaKey) layer that modulates motion latents for keyframe-conditioned synthesis while preserving the original latent manifold. To further adapt the model for \emph{scene-aware} inbetweening, we introduce a \emph{Scene-conditioning} (SceneCo) layer that injects geometric scene information by adaptively querying local context via cross-attention. Experimental results show that \textbf{SceneAdapt} effectively injects scene-awareness into text-to-motion models without sacrificing semantic diversity, and we further analyze the mechanisms through which this awareness emerges. Code and models will be released. 
\textbf{Project page}: \href{https://sceneadapt.github.io/}{sceneadapt.github.io}

\end{abstract}

\section{Introduction}

Generating realistic human motion has attracted significant attention, with broad applications in virtual reality, gaming, and robotics. 
For practical use, motion models must satisfy two goals: achieving the \emph{semantic richness and naturalness} of everyday actions, and ensuring \emph{physical consistency} with the surrounding scene.
Failing the former yields motions that are incoherent, while neglecting the latter produces physically implausible results, such as walking through walls. 
Existing approaches, however, fail to jointly ensure the semantic diversity of motion and its consistency with the scene.

\begin{figure}[t]
    \centering
    \includegraphics[width=0.95\textwidth]
    {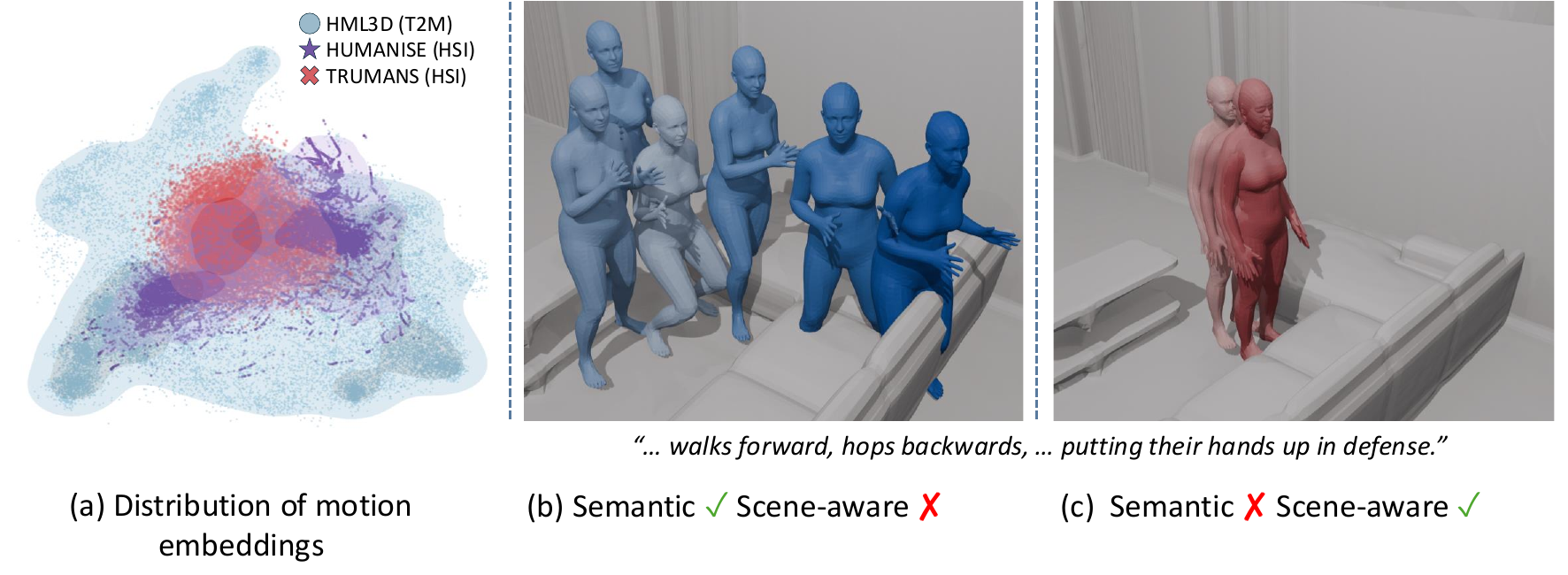}
    \caption{\textbf{Motivation.} (a) Distribution of motion embeddings of HML3D~\cite{humanml3d} and scene-aware datasets~\cite{humanise, jiang2024trumans} visualized via PCA. Scene-aware datasets show narrower distributions than HML3D, indicating lower semantic coverage. 
    (b) Models trained on HML3D capture diverse action semantics but lack scene-awareness, penetrating the obstacles. (c) Models trained on scene-aware datasets satisfy scene constraints, but fail to follow text prompts because the datasets contain limited semantic motion diversity.}
    
        \vspace{-10pt} 
    \label{fig:motivation}
\end{figure}

On the semantic side, text-conditioned motion models \cite{guy2023mdm, chen2023mld}, trained on large-scale text–motion corpora \cite{babel, humanml3d, snapmogen}, can synthesize semantically rich motions from language, showing strong generalization to diverse text prompts.
Yet, as these models only target text-to-motion, they remain blind to spatial constraints, generating motions that do not adhere to the given scene (Fig.~\ref{fig:motivation}.b).


On the other hand, scene-aware motion generation aims to synthesize motions that satisfy physical constraints within the surrounding scene (e.g., collision avoidance), while remaining aligned with additional signals such as text. However, capturing motion with precise scene context typically requires professional MoCap systems, whose high cost prevents scaling to semantically diverse scenarios. 
As a result, early works~\cite{humanise,HSI_dataset,circle} relied on synthetic data, and even recent motion capture datasets~\cite{jiang2024trumans} remain limited to a narrow set of everyday actions (e.g., walking, sitting) as shown in Fig.~\ref{fig:motivation}.a.
Consequently, models trained on these datasets cannot generalize beyond restricted actions (Fig.~\ref{fig:motivation}.c).

Motivated by existing limitations, we are interested in developing a model capable of synthesizing motions that are both semantically rich and scene-aware. For example, generating “a person walking in a circle’’ or “a man dribbling a basketball’’ in a 3D scene requires understanding both motion semantics and geometric scene constraints.
However, collecting large-scale text–scene–motion datasets is infeasible. Therefore, we ask: \emph{can we generate scene-aware and semantically rich motion from text by leveraging existing, but disjoint, datasets?}

In this paper, we introduce \textbf{SceneAdapt}, a two-stage adaptation framework that injects scene-awareness into a pretrained motion diffusion model (MDM), using only existing text-motion and scene-motion datasets.
Our key insight is to leverage motion inbetweening,
which can be learned without text,
as a proxy task to inject scene-awareness and enable scene-aware text-conditioned generation.
To be specific, we first adapt MDM to motion inbetweening, and then further adapt it for scene-aware inbetweening~\cite{scenemi} using only scene–motion pairs.
Since the model already learns inbetweening in the first stage, the second stage focuses exclusively on leveraging scene data to achieve scene-consistent inbetweening, thereby injecting scene-awareness into the model.

To adapt text-conditioned models for inbetweening, we design a Context-aware Keyframing (CaKey) layer that selectively modulates keyframe latents, enabling accurate inbetweening while preserving the pretrained latent manifold of the text-to-motion model.
In the second stage, we freeze the CaKey layer and train only the Scene-conditioning (SceneCo) layers which use cross-attention to inject scene-awareness.
Whereas prior works \cite{jiang2024trumans} use global features, we utilize patch-wise features, allowing frame-wise latents to focus on different places in the scene.
Through these adaptations, the model can generate motions that are both faithful to text prompts and consistent with the surrounding scene. 

Extensive experiments demonstrate that 
\textbf{SceneAdapt} genuinely exploits scene information, leading to motions that are both semantically rich and scene-aware.
Furthermore, we show that the proposed components at each stage lead to significant performance gain, validating our overall pipeline as effective.
We further analyze how scene-awareness is injected into the model, providing new insights into the mechanisms through which text-conditioned motion generation benefits from scene information. 

The main contributions of this work are:
(1) We propose \textbf{SceneAdapt}, a two-stage adaptation framework that injects scene-awareness into a pretrained motion diffusion model using only text-motion and scene-motion datasets. 
(2) We design \textbf{Context-aware Keyframing (CaKey) layer}, which modulates only keyframe latents to enable faithful motion inbetweening without distorting the original text-to-motion manifold.
(3) We introduce a \textbf{Scene-conditioning (SceneCo) layer} that leverages cross-attention between frame-wise motion latents and voxel patch features to inject scene-awareness. %
(4) Extensive experiments show that \textbf{SceneAdapt} outperforms from-scratch baselines, improves scene-awareness in text-to-motion generation, and provides insights into how scene information is integrated into generative models.

\section{Related Work}
\vspace{-2pt}
\paragraph{\textbf{Text-to-motion (T2M) synthesis.}} 
Given a text description, text-to-motion (T2M) generation aims to generate corresponding natural and diverse motions.
Early works employed models such as RNNs or Transformers~\cite{action2motion, petrovich2022temos, zhang2023t2m, siyao2022bailando, t2mgpt}, and focused on alignment between motion and language latent spaces~\cite{ahuja2019language2pose, tevet2022motionclip}. Recently, \cite{guy2023mdm} introduced the Motion Diffusion Model (MDM), a text-conditioned motion generator trained on large-scale text–motion datasets~\cite{KITdataset2016, humanml3d}, demonstrating strong generative performance. Subsequent works~\cite{zhong2023attt2m, chen2023mld, motionLCM, pinyoanuntapong2024mmm, pinyoanuntapong2024bamm, FlowMDM, momask, emdm, physdiff,zhang2025motionanythingmotiongeneration, primal, zhang2025flashmo, li2025lamp, Cho_2025_ICCV} have further improved generation quality, efficiency, semantic alignment, or physical plausibility. However, as these datasets lack scene context, the resulting models remain inherently unaware of their surroundings.
\vspace{-2pt}
\paragraph{\textbf{Scene-aware T2M synthesis.}}
Scene-aware text-to-motion~\cite{cong2024laserhuman, yi2024tesmo, cen2024text_scene_motion, affordmotion, humanise, jiang2024trumans, huang2023diffusion} aims to generate motions that are not only natural and faithful to the textual description but also physically consistent with a 3D scene. 
However, obtaining real motion data that is accurately aligned with surrounding scene geometry remains extremely challenging. To address this limitation, several recent works~\cite{black2023bedlam, circle, humanise, yi2024tesmo, cen2024text_scene_motion} construct synthetic scene–motion datasets as a scalable alternative to expensive real-world capture.
For instance, HUMANISE~\cite{humanise} introduced a large-scale synthetic dataset by aligning the scanned indoor scenes with captured motion sequences.
Although such datasets enable scalable training as done in ~\cite{affordmotion}, they fall short in capturing the realism of actual human–scene interactions. 
Recent works~\cite{jiang2024trumans, lingo, zhang2024scenic, circle, cong2024laserhuman} have proposed real-world MoCap datasets captured with professional apparatus.
However, these datasets remain impractical due to their limited motion semantic diversity--only containing simple motions such as walking, sitting down, picking, or standing up--and are difficult to scale due to the high cost of capturing both motion and 3D scene.
To avoid reliance on datasets, some studies~\cite{li2023genZI, li2025zeroHSI} leverage pretrained image or video diffusion models for zero-shot motion generation, but struggle to generate realistic motions. In light of these limitations, throughout this paper we define scene-aware T2M synthesis as generating semantically diverse motions that are physically consistent with a given 3D scene. Accordingly, instead of collecting new datasets or relying on indirect solutions, we leverage existing motion corpora and introduce a novel adaptation strategy that enables semantically rich motions while adhering to the given scene constraints. Detailed comparison with prior works are provided in the supplementary materials. (Suppl.\S~{\color{red}{10}})

\vspace{-2pt}
\paragraph{\textbf{Spatially controlled T2M synthesis.}} 
Several studies~\cite{gmd, zhao2025dartcontrol, hoidini, coda} have focused on spatial control by propagating gradients from external conditions, such as pelvis trajectories, 2D obstacles, or even objects, into the initial diffusion noise~\cite{karunratanakul2024dno}.
However, these methods require hundreds of optimization steps, leading to extreamly slow synthesis.
Moreover, they often fail to reflect the text descriptions, as satisfying scene constraints takes priority.
In contrast, we present a feed-forward approach that generates motions that are both scene-aware and faithful to text conditions.

\vspace{-2pt}
\paragraph{\textbf{Adaptation of Diffusion Models.}} Diffusion models pretrained on large-scale datasets~\cite{latentdiffusionmodel, peebles2023scalablediffusionmodelstransformers} demonstrate impressive generative ability, but often require adaptation to new conditions or domains.
One representative method is ControlNet~\cite{zhang2023controlNet}, which augments a frozen network with a trainable copy, enabling generation guided by various signals such as pose, edge, or depth maps.
Another widely used strategy is LoRA~\cite{hu2021lora}, which adapts pretrained diffusion models to novel domains in a parameter-efficient manner~\cite{guo2024animatediff, shi2024dragdiffusion}.
Recent efforts introduce auxiliary modules to incorporate additional control signals such as camera parameters~\cite{he2025cameractrl, wang2024motionctr, xu2024camco, bai2025recammaster}, or user action controls~\cite{yu2025gamefactory}.
Among these, \cite{yu2025gamefactory} introduces a multi-phase adaptation pipeline, which motivates our strategy.
While \cite{yu2025gamefactory} adapts a video diffusion model to respond to interactive controls like keyboard inputs, our method instead equips a text-conditioned motion diffusion model with 3D scene-awareness.

\section{Method}

{\setlength{\textfloatsep}{0pt} 
 \setlength{\intextsep}{0pt}
\begin{figure}[t]
    \centering
    \includegraphics[width=0.96\textwidth]{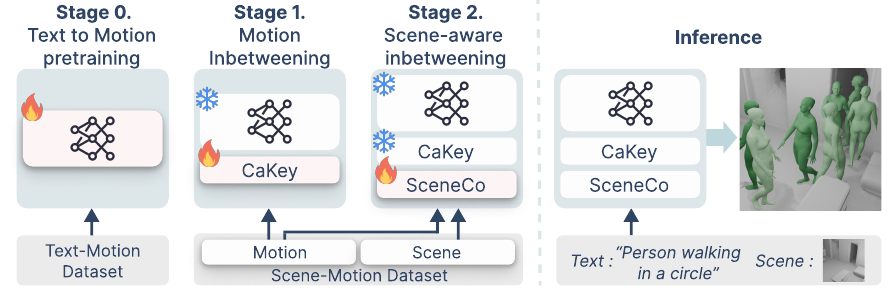}
    \captionsetup{skip=2pt}
\caption{\textbf{Overview.} Starting from a pretrained text-to-motion model (\textbf{Stage 0}), we first insert CaKey layers and train them with a motion inbetweening objective (\textbf{Stage 1}), which only requires motion sequences. We then add scene-conditioning (SceneCo) layers and train them with a scene-aware inbetweening objective (\textbf{Stage 2}), using scene-motion pairs. During inference, we use the base model and adaptors to generate semantically rich motion which also adheres to the scene geometry.}
    \label{fig:main_method}
\end{figure}
\vspace{-3mm} 
}
The overall pipeline of SceneAdapt is illustrated in Fig.~\ref{fig:main_method}.
Given a pretrained T2M model (\textbf{Stage 0}), we first adapt it for \textit{motion inbetweening} (\textbf{Stage 1}, \S~{\color{red}\ref{sec:inbetween}}) using our novel CaKey layers, which generates natural motions consistent with input keyframes.
Next, we freeze the CaKey layers and insert scene-conditioning layers (SceneCo) to learn scene-aware inbetweening (\textbf{Stage 2}, \S~{\color{red}\ref{sec:scene-inbetween}}).
At inference, we use the trained adapters to perform scene-aware text-to-motion generation (\S~{\color{red}\ref{sec:text-scene-motion}}). 
For implementation details, see Suppl.\S~{\color{red}{3}}.

.
\vspace{-6pt} 
\subsection{Preliminaries}
\vspace{-6pt} 
\paragraph{\textbf{Problem Formulation. }}
We define a 3D scene as $\mathcal{S}$, a text prompt as $\mathcal{T}$, and a keyframe mask $m^{1:N} = \{m^i\}_{i=1}^N$ with $m^i \in \{0,1\}$, where $m^i = 1$ indicates that the $i^{th}$ frame is a keyframe.
Our goal is to generate a natural motion sequence $x^{1:N} = \{x^i\}_{i=1}^{N}$, where $x^i \in \mathbb{R}^D$, conditioned on different forms of context:
(i) \textit{motion inbetweening}, which models $p({x^{1:N}}\mid{m^{1:N}})$;  
(ii) \textit{scene-aware inbetweening}, which models $p({x^{1:N}}\mid{m^{1:N}}, \mathcal{S})$;  
(iii) \textit{scene-aware text-conditioned generation}, which models $p({x^{1:N}}\mid {\mathcal{S}, \mathcal{T}})$.



\paragraph{\textbf{Motion Representation. }} We adopt the HML3D~\cite{humanml3d} representation, where each pose $x^i$ is a 263-dimensional vector. Following~\cite{conmdi}, we convert the relative root orientation and the relative $x,z$ positions into their global counterparts, which allows us to adapt MDM for motion inbetweening. Additional details are provided in the supplementary materials. (Suppl.\S~{\color{red}{3}})


\paragraph{\textbf{Motion Diffusion Model. }} We adopt MDM~\cite{guy2023mdm} as our baseline model, following prior works~\cite{loramdm, omnicontrol, karunratanakul2024dno, yi2024tesmo}. 
MDM models text-conditioned motion generation within the DDPM framework \cite{ddpm}, which consists of a forward and a backward diffusion process. 
The forward diffusion is formulated as a Markov noising process that produces a sequence $\{x_t\}_{t=0}^T$, 
where $x_0$ is the clean data and $t$ is the diffusion timestep. 
Each step is defined as
$q({x_t}\mid{x_{t-1}}) = \mathcal{N}(x_t; \sqrt{1-\beta_t}\,x_{t-1}, \beta_t \mathbf{I})$,
with $\{\beta_t\}_{t=1}^T$ denoting the variance schedule.
During the backward pass, instead of predicting the noise $\epsilon$, the denoising network is parameterized to directly predict the clean motion 
$\hat{x}_0 = \mathcal{D}_\theta(x_t, t, \mathcal{T})$.
The training objective is the simplified $L_2$ reconstruction loss,
\begin{equation}\label{equation:diffusion_loss}
\mathcal{L}_{\text{t2m}} = 
\mathbb{E}_{x_0 \sim q(x_0 \mid \mathcal{T}),\, t \sim [1,T]}
\Big[ \big\| x_0 - \mathcal{D}_\theta(x_t, t, \mathcal{T}) \big\|_2^2 \Big].
\end{equation}
with additional geometric losses applied in the raw motion space.

\subsection{Stage 1: Adaptation for Inbetweening}
\label{sec:inbetween}
Prior approaches have explored inbetweening either by imputing keyframes at inference time~\cite{guy2023mdm} or by training specialized models from scratch \cite{conmdi, scenemi}. However, the adaptation of text-conditioned motion generation models to the inbetweening setting remains unexplored. A well-adapted inbetweening model should not only achieve high keyframe alignment, but also preserve the naturalness and text-adherence capability of the original model.

\paragraph{\textbf{Adaptation Layers.}} To achieve these properties, we introduce the Context-aware Keyframing (CaKey) layer,
which applies affine modulation to the MDM latents based on the given keyframes. CaKey introduces two key modifications over standard FiLM-style modulation:
(1) \textbf{Context-awareness}. The modulation parameters are estimated not only from the keyframe signal but also from the diffusion timestep along with the latent representation being modulated, enabling the modulation to be aware of \textit{what it is modulating}, and thereby improving alignment with input keyframes.
(2) \textbf{Sparse modulation}. Identity is preserved on non-keyframe indices while modulation is applied only on the keyframe indices, ensuring that only the keyframe latents are modulated, thereby preserving the original generation capabilities of the base model.



\setlength{\intextsep}{0pt}        
\setlength{\columnsep}{5pt}       

\begin{wrapfigure}{r}{0.45\linewidth}
  \vspace{-2pt} 
  \centering
  \includegraphics[width=\linewidth]{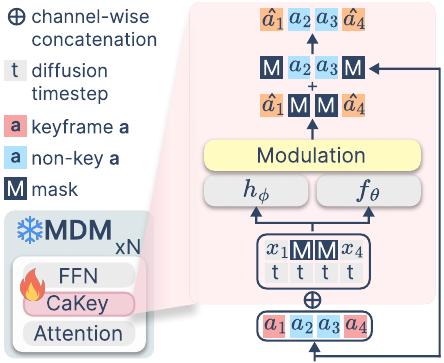}
  \captionsetup{skip=2pt} 
  \caption{\textbf{CaKey Layer.} We use adaptive modulation layers.}
  \vspace{-2pt} 
  \label{fig:caky}
\end{wrapfigure}

Formally, CaKey employs two learnable MLP-based networks, $f_\theta$ and $h_\phi$.
These networks take as input the ground-truth motion $x$, the diffusion timestep $t$, and the current self-attention activation $a$, and output the scale $\gamma$ and shift $\beta$ parameters:
\begin{equation}
\gamma = f_\theta(x, t, a),\quad \beta = h_\phi(x, t, a)
\end{equation}
Our modulation process is described as
\begin{equation}
\hat{a} = \gamma \odot a + \beta,
\end{equation}
\begin{equation}
\text{CaKey}(a, m, x, t) = (1-m)\odot a + m \odot \hat{a},
\end{equation}
where $\odot$ denotes element-wise multiplication and $m$ denotes keyframe mask.



\paragraph{\textbf{Training.}}
We freeze the base MDM parameters and optimize only the CaKey layers under the motion inbetweening objective. 
The loss follows the diffusion formulation in Eq.~\ref{equation:diffusion_loss}, with two modifications: (i) the text input is replaced by the null embedding $\varnothing_{text}$, and (ii) conditioning is augmented with a keyframe mask $m^{1:N}$. The mask is sampled randomly with a fixed stride $s_k$, while the first and last frames are always designated as keyframes ($m^{0} = m^{N} = 1$). In all experiments in stage 1, we set $s_k=20$ which corresponds to one keyframe per second.



\subsection{Stage 2: Adaptation for Scene-aware Inbetweening}
Building on stage 1, we introduce Scene-conditioning (SceneCo) layers and train them while keeping the rest of the model frozen.
It is worth emphasizing that the learned inbetweening capability at stage 1 allows scene-aware learning to become the primary objective in minimizing the training loss.
This design thus encourages the new parameters to focus solely on leveraging scene information, thereby injecting scene-awareness into the model.

\label{sec:scene-inbetween}

\paragraph{\textbf{Scene Representation.}}
We voxelize the scene $\mathcal{S}$ into a binary occupancy grid $\mathcal{V} = \text{voxelize}(\mathcal{S}) \in \{0,1\}^{d_x \times d_y \times d_z}$, where $1$ denotes an occupied cell and $0$ a free one. Previous approaches encode $\mathcal{V}$ into a single global vector via the class embedding of a Voxel ViT~\cite{jiang2024trumans, lingo}, conditioning all frames on the same vector~\cite{scenemi}. 
However, such global features overlook the fact that \textit{joint positions evolve over time, and thus different frames interact with different local neighborhoods of the scene}.
To capture this spatio-temporal variation, we instead use the patch embeddings from a voxel ViT and then let SceneCo layers interact between motion latents and voxel patch embeddings.
Specifically, we obtain patch embeddings from the voxel ViT: $s = \text{ViT}(\mathcal{V}) \in \mathbb{R}^{P \times d_s},$
where $P$ is the number of spatial patches and $d_s$ the embedding dimension.


\paragraph{\textbf{Adaptation Layers.}} 
We employ Scene-conditioning (SceneCo) layers, which are simple cross-attention layers, where motion latents query voxel patches so that each frame can selectively attend to its relevant local context. 
Formally, let $h = \{h^i\}_{i=1}^{N+1} \in \mathbb{R}^{(N+1) \times d}$ denote the latent sequence, where $h^1$ corresponds to the text token and $\{h^i\}_{i=2}^{N+1}$ to motion frames. Let $s = \{s^j\}_{j=1}^{p_n} \in \mathbb{R}^{p_n \times s_{dim}}$ be the patch embeddings of the voxelized scene. Cross-attention is then defined as
\[
h_{out} = \text{ATT}(hW_Q, sW_K, sW_V)
\]

To ensure that scene information is used only where necessary, we mask activations as follows: (i) \textbf{the text token $h^1$} and (ii) padded frames,
leaving scene conditioning active only for 
motion latents 
that require scene-awareness.


\paragraph{\textbf{Training.}} We keep the MDM and CaKey layers frozen, and train the additional cross-attention layers along with our voxel ViT on the motion inbetweening objective using 3D scenes as inputs.
A key challenge at this stage is that, unlike the previous stage where only keyframes are modulated, the cross-attention layers broadly affect the motion latent space, leading to a decline in the model’s original text-to-motion performance. 
To mitigate this issue, we utilize the text-motion paired dataset used during pretraining for prior preservation~\cite{dreambooth, loramdm}, by adding Eq.~\ref{equation:diffusion_loss}. 
As text-motion paired datasets do not provide 3D scenes, we introduce a learnable null embedding $\varnothing_{scene}$ for the \textbf{prior loss}, while dropping 10$\%$ of text inputs for classifier-free guidance.
We also apply $\varnothing_{scene}$ for 10$\%$ of the scene features in the scene-motion pairs.

\subsection{Text to Scene-aware Motion Generation} \label{sec:text-scene-motion}
With both CaKey layers and SceneCo layers trained, we perform scene-aware text-conditioned motion generation by conditioning the final model only on text and scene inputs, while using an all-zero keyframe mask ($m^{1:N} = 0$), indicating that no keyframes are provided. 

\paragraph{\textbf{Sampling.}} We introduce two classifier-free guidance scales: $w_t$ for text guidance and $w_s$ for scene guidance. These scales control the trade-off between semantic alignment with the text and physical consistency with the scene during motion generation.  Formally,
\begin{equation}
\begin{split}
    \hat{x}_0 &= \mathcal{D}_\theta(x_t, t, \varnothing_{text}, \varnothing_{scene}) 
    + w_t \big( \mathcal{D}_\theta(x_t, t, \mathcal{T}, \varnothing_{scene}) - \mathcal{D}_\theta(x_t, t, \varnothing_{text}, \varnothing_{scene}) \big) \\
    &+ w_s \big( \mathcal{D}_\theta(x_t, t, \varnothing_{text}, \mathcal{S}) - \mathcal{D}_\theta(x_t, t, \varnothing_{text}, \varnothing_{scene}) \big).
\end{split}
\end{equation}

\paragraph{\textbf{Goal pose conditioning.}} Our objective is to generate motions that are semantically rich and scene-aware (e.g., avoiding penetration into the environment). However, since our approach does not model scene–semantic relationships (e.g., “walk to the refrigerator”), it cannot directly produce functional behaviors that require semantic understanding of the scene. Nevertheless, since our adaptation method is based on sparse keyframing, we can achieve goal-directed, scene-aware text-to-motion~\cite{closd, lingo, zhao2025dartcontrol} by additionally conditioning the model on goal poses. In this setting, the model is able to sit on chairs and reach toward objects across diverse scenes when provided with a goal pose, the scene, and a text description. 
The results are shown in Figure~\ref{fig:goal_conditioned_qaulitative}. 
Leveraging scene-aware pose generation methods to further guide SceneAdapt toward functional, goal-directed motion represents a promising direction for future work.


\section{Experiments}
We first evaluate SceneAdapt on scene-aware text-conditioned motion generation (\S~{\color{red}\ref{eval:sa_t2m}}), then assess the effectiveness of CaKey for motion inbetweening, and further examine how incorporating scene-conditioning layers injects scene-awareness (\S~{\color{red}\ref{eval:mi}}). 
Finally, we conduct a component-wise ablation study to validate the contribution of each design choice (\S~{\color{red}\ref{eval:analysis}}).

\paragraph{\textbf{Training Dataset.}} The baseline MDM model is trained on the text–motion paired HML3D dataset~\cite{humanml3d}.
For adaptation, we additionally use the scene–motion paired TRUMANS dataset~\cite{jiang2024trumans}, a high-quality mocap dataset with precise alignment to scene geometry. 
While HML3D is represented using the skeleton from SMPL-H~\cite{smplh}, TRUMANS is provided in SMPL-X~\cite{smplx}. 
To ensure compatibility, we fit SMPL-H meshes to TRUMANS motions and follow the preprocessing pipeline of~\cite{humanml3d}.
TRUMANS sequences are relatively slow and long, recorded at 30 FPS. We downsample them by a factor of 2 (15 FPS) and segment them into 196-frame clips.
Although HML3D is at 20 FPS, the slower dynamics of TRUMANS make the downsampled sequences match the speed of HML3D.
\paragraph{\textbf{Evaluation Metrics.} } We compute Frechet Inception Distance (\textbf{FID}) to measure the overall diversity and naturalness of the generated motions and R-Precision (\textbf{RP}) \cite{humanml3d} to evaluate the text-adherence to the given prompt. 
Motivated by prior scene-aware works~\cite{zhang2020generating3dpeoplescenes, scenemi, humanise}, we holistically assess geometry compliance using three metrics.
Collision-frame ratio (\textbf{CFR}) measures \emph{how often} violations occur: the fraction of frames with any penetration.
Mean max penetration (\textbf{MMP}) measures \emph{how severe} a violation is when it happens: the average per-frame deepest penetration (m) over colliding frames.
Joint-collision ratio (\textbf{JCR}) measures \emph{how widespread} a violation is: the mean fraction of joints penetrating, computed \emph{only over colliding frames} (pure extent), thus decoupled from CFR's frequency. 
We define penetration using signed distance fields (SDFs) with a 2\,cm tolerance: letting $d_{t,v}$ be the signed distance of joint $v$ at frame $t$ (negative inside), a joint is counted as colliding iff $d_{t,v}<-\delta$ with $\delta{=}2$\,cm.
For inbetweening, we quantify the mean joint position error (\textbf{MJPE}) for both the full sequence and the keyframes to measure keyframe alignment. 
We further report \textbf{foot skating}~\cite{gmd} and \textbf{skating ratio}~\cite{zhang2022wanderings} to quantify sliding artifacts.

\paragraph{\textbf{Evaluation Dataset.}} Evaluating scene-aware text-to-motion generation with diverse semantics requires a test set that is (1) aligned with 3D scenes (for CFR, MMP, JCR) and (2) has a large amount of diverse motion with different semantics (for FID and RP). However, since \textit{text–scene–motion triplet dataset with diverse motion semantics does not exist}, we augment the HML3D test set by matching each motion-text pair with a sampled trajectory position and rotation from TRUMANS motion sequences, which serve as the first frame’s global position and rotation. To avoid implausible setups, we sample only from TRUMANS frames with sufficient clearance from surrounding geometry and exclude HML3D texts that require specific terrain (e.g., stairs or climbing) not present in TRUMANS.
This yields text–scene–motion pairs that enable us to evaluate text-to-motion generation using model-based metrics (FID and RP) as well as our scene-aware metrics (CFR, MMP, JCR). 
Details of this evaluation set construction are provided in the supplementary materials (Suppl.\S~{\color{red}{1}}).

\paragraph{\textbf{Baselines.}} For scene-aware text-to-motion generation, we compare SceneAdapt against two groups of state-of-the-art baselines. (1) Optimization-based methods trained on HML3D with text-motion pairs; DNO~\cite{karunratanakul2024dno} and DART~\cite{zhao2025dartcontrol}. (2) Feed-forward baselines trained on synthetic text–scene–motion triplets from HUMANISE~\cite{humanise} (and additionally text-motion pairs on HML3D); MDM augmented with additional modules, HUMANISE cVAE and AffordMotion~\cite{affordmotion}. Note that these models cannot be trained with TRUMANS due to the lack of text prompts and diversity of motions. \footnote{While open-sourced text descriptions are available, they are brief action tags (e.g., \emph{sit up}, \emph{sit down}) and lack semantically rich motion descriptions.}
For motion inbetweening, we benchmark against imputation-based sampling~\cite{guy2023mdm}, MDM augmented with LoRA~\cite{hu2021lora}, and CondMDI~\cite{conmdi} which is specifically designed for inbetweening.

\begin{table*}[t]
\centering
\small
\setlength{\tabcolsep}{4.2pt} 
\resizebox{\textwidth}{!}{%
\begin{tabular}{l|lccccccccc}
\toprule
\textbf{Group} & \textbf{Method} &
\textbf{\hspace{1pt}T--M\hspace{1pt}} & \textbf{\hspace{1pt}T--S--M\hspace{1pt}} & \textbf{\hspace{1pt}S--M\hspace{1pt}} &
\textbf{RP@3$\uparrow$} & \textbf{FID$\downarrow$} &
\textbf{CFR$\downarrow$} & \textbf{MMP$\downarrow$} & \textbf{JCR$\downarrow$} & \textbf{Inf. Time (s)$\downarrow$} \\
\midrule

\multirow{1}{*}{\textbf{\shortstack{Baseline}}}
& MDM
& \cmark & \xmark & \xmark
& 0.798 & 0.479 & 0.316 & 0.319 & 0.344 & 0.52 \\
\midrule

\multirow{5}{*}{\textbf{\shortstack{Feed-\\forward}}}
& MDM + ControlNet
& \cmark & \cmark & \xmark
& 0.365 & 36.19 & 0.142 & 0.041 & 0.064 & 1.79 \\
& MDM + SceneCo Layer
& \cmark & \cmark & \xmark
& 0.094 & 80.89 & 0.050 & 0.004 & 0.005 & 1.69 \\
& HUMANISE cVAE
& \xmark & \cmark & \xmark
& 0.092 & 34.58 & 0.002 & 0.001 & 0.001 & 0.17 \\
& AffordMotion
& \xmark & \cmark & \xmark
& 0.140 & 21.59 & 0.257 & 0.059 & 0.097 & 50.72 \\
& AffordMotion
& \cmark & \cmark & \xmark
& 0.305 & 6.320 & 0.429 & 0.254 & 0.321 & 51.28 \\
\midrule

\multirow{2}{*}{\textbf{\shortstack{Optimi-\\zation}}}
& DNO
& \cmark & \xmark & \xmark
& 0.128 & 32.22 & 0.001 & 0.002 & 0.002 & 332.96 \\
& DARTControl
& \cmark & \xmark & \xmark
& 0.056 & 53.29 & 0.010 & 0.007 & 0.010 & 362.90 \\
\midrule

\rowcolor{gray!15} 
\multirow{5}{*}{\cellcolor{white}\textbf{\shortstack{Ours}}}
& Ours ($w_s$ = 0.3)
& \cmark & \xmark & \cmark
& 0.792 & 0.497 & 0.256 & 0.208 & 0.246 &  \\
& Ours ($w_s$ = 0.0)
& \cmark & \xmark & \cmark
& 0.803 & 0.312 & 0.298 & 0.273 & 0.299 &  \\
& Ours ($w_s$ = 0.5)
& \cmark & \xmark & \cmark
& 0.750 & 1.420 & 0.220 & 0.160 & 0.199 & 1.69 \\
& Ours ($w_s$ = 1.0)
& \cmark & \xmark & \cmark
& 0.588 & 7.389 & 0.136 & 0.076 & 0.101 &  \\
& Ours ($w_s$ = 2.0)
& \cmark & \xmark & \cmark
& 0.365 & 18.88 & 0.072 & 0.035 & 0.045 &  \\
\bottomrule
\end{tabular}}

\caption{\textbf{Scene-aware text-driven generation results on our evaluation set}. 
T--M indicates training with text--motion pairs (e.g., HML3D), T--S--M indicates text--scene--motion triplets (e.g., HUMANISE), and S--M indicates scene--motion pairs (e.g., TRUMANS). 
“Inf. Time” reports the average inference time per sample in RTX A5000.}
\vspace{-20pt}
\label{tab:sa_t2m}
\end{table*}
\vspace{-10pt}
\subsection{Scene-aware text conditioned motion generation} \label{eval:sa_t2m}

\paragraph{\textbf{Quantitative results.}} As shown in Tab.~\ref{tab:sa_t2m}, compared to MDM, our adaptation improves MDM's scene-awareness without sacrificing its text-to-motion capabilities. The two variants that equip MDM with additional scene-conditioning modules fail to preserve MDM’s text--motion alignment and motion naturalness during adaptation. Compared to HUMANISE cVAE and AffordMotion, our method achieves superior performance in both text-to-motion alignment and scene-awareness (see \textit{Ours} ($w_s=0.5$)), showing that training solely on high-quality scene--motion pairs can outperform models trained with synthetic text--scene--motion triplets of limited semantic coverage. 
While optimization-based methods can achieve nearly perfect scene-awareness by directly optimizing motions to avoid penetrating 3D scenes, they often fail to preserve the original T2M model’s generative capabilities and incur slow inference. 
In contrast, SceneAdapt preserves the T2M model’s generative strengths while also delivering strong scene-awareness. 
Overall, SceneAdapt combines high scene-awareness with strong text alignment and naturalness, while remaining orders of magnitude faster than optimization-based baselines, making it a practical solution for scalable scene-aware motion generation.
To provide a more detailed view of scene-awareness performance, we report penetration-depth distributions and summary statistics in the supplementary materials (Suppl.\S~{\color{red}{4}}).


\begin{figure}[t]
    \centering
    \includegraphics[width=0.99\textwidth]{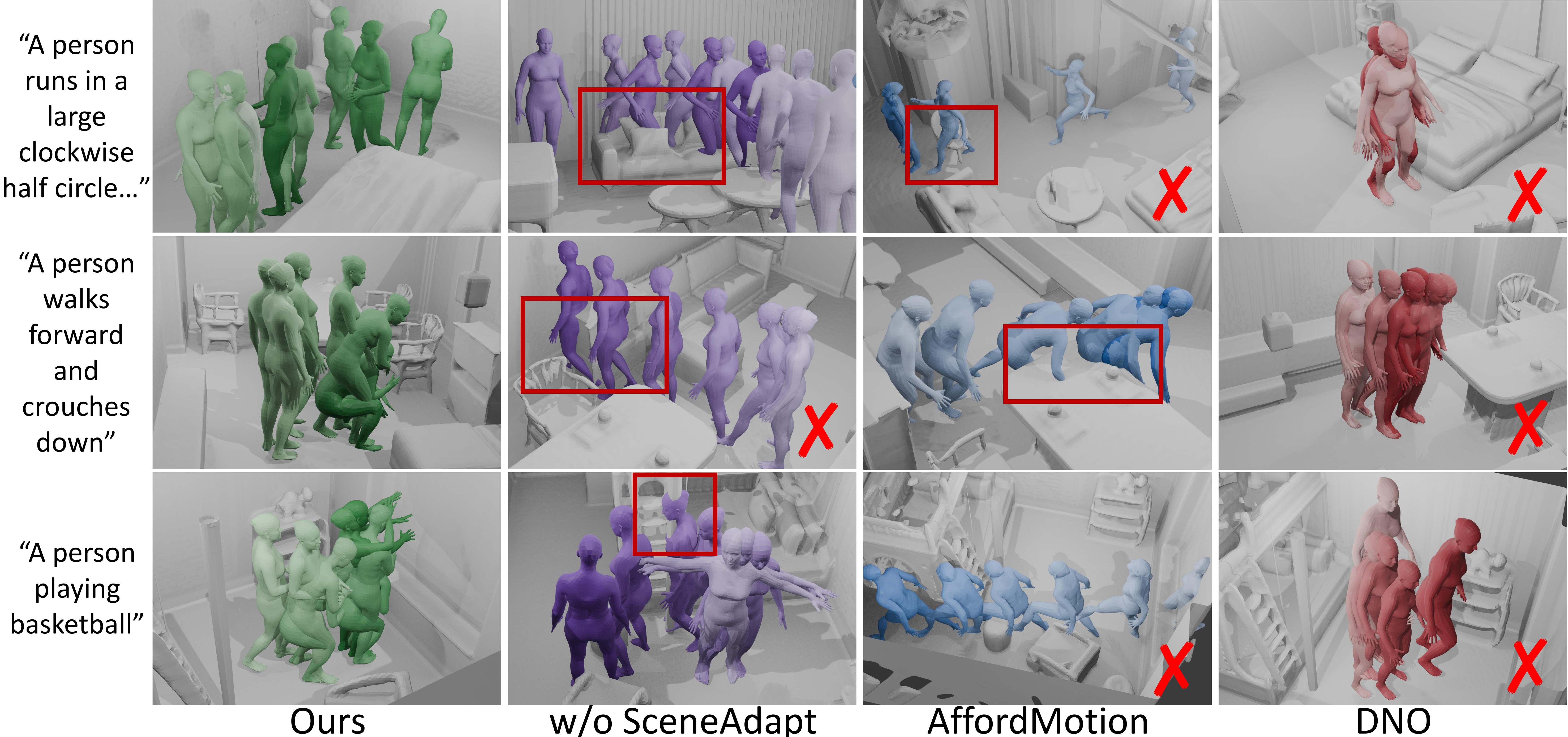}
    \vspace{-5pt} 
    \caption{\textbf{Qualitative results.}  {\color{red}Boxes} mark collisions and {\color{red}X}s mark semantic errors. Unlike AffordMotion (scene penetration) and DNO (weak text alignment), our method improves MDM by enhancing scene-awareness while preserving text fidelity.}
    \label{fig:qaulitative}
    \vspace{-5pt} 
\end{figure}

\begin{figure}[t]
    \centering
    \includegraphics[width=0.99\textwidth]{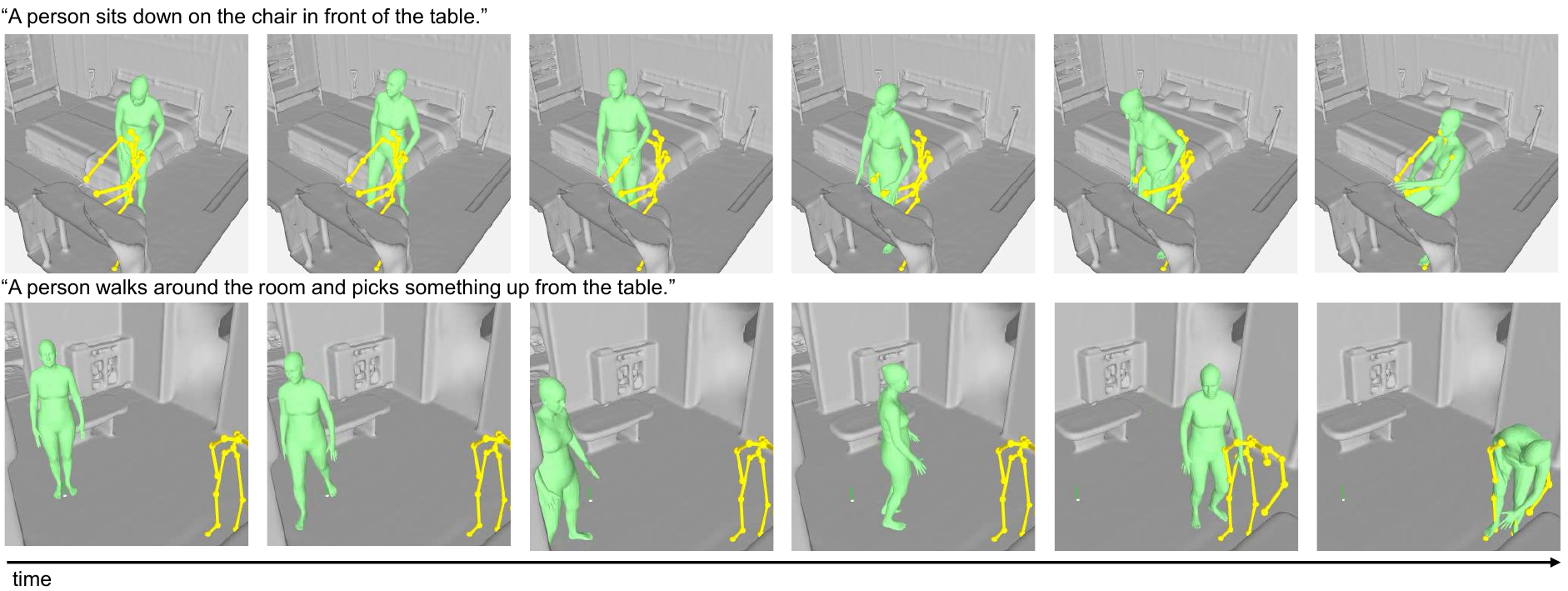}
    \vspace{-10pt} 
\caption{\textbf{Goal pose conditioned scene-aware text-to-motion generation.}
Interpreting the goal pose as an extremely sparse keyframe, SceneAdapt produces scene-consistent motion conditioned on text, scene, and goal pose. Goal poses are in yellow.}
    \label{fig:goal_conditioned_qaulitative}
    \vspace{-15pt} 
\end{figure}

\paragraph{\textbf{Qualitative results.}} As shown in Fig.~\ref{fig:qaulitative}, AffordMotion suffers from scene penetration or weak adherence to text prompts, reflecting limitations of the HUMANISE dataset, which contains only synthetic scene–motion interactions with limited diversity. DNO achieves strong scene-awareness but sacrifices text alignment during the optimization process. In contrast, our method equips MDM with scene-awareness, substantially reducing scene penetration while preserving text fidelity. Moreover, in Fig.~\ref{fig:goal_conditioned_qaulitative}, we show that conditioning on an additional goal pose enables our model to generate motions that interact with the scene without penetrating it, while still following the text prompt.
See the supplementary materials for additional results.


\begin{table*}[t]
  \centering

  \setlength{\abovecaptionskip}{6pt}   
  \setlength{\belowcaptionskip}{0pt}   

  \setlength{\tabcolsep}{1pt}
  \renewcommand{\arraystretch}{1.2}

  \resizebox{\textwidth}{!}{%
    \begin{tabular}{lcccccc}
      \toprule
      \textbf{Method} & \textbf{RP(top 3)}$\uparrow$ & \textbf{FID}$\downarrow$ &
      \textbf{MJPE(Key)}$\downarrow$ & \textbf{MJPE(All)}$\downarrow$ &
      \textbf{Foot skating}$\downarrow$ & \textbf{Skating ratio}$\downarrow$ \\
      \midrule
GT & 0.7980 & 0.002 & 0 & 0 & - & - \\
\midrule
MDM (imputation) & 0.6144 & 7.258 & 0 & 0.7647 & 0.1012 & 0.3971 \\
MDM + LoRA & 0.7214 & 0.074 & 0.0625 & 0.1120 & \textbf{0.0418} & 0.0625 \\
CondMDI & 0.6767 & 0.356 & 0.2804 & 0.2957 & 0.1067 & 0.1074 \\
\midrule
\rowcolor{gray!15}
\cellcolor{white}{Ours} & \textbf{0.7242} & \textbf{0.036} & 0.0018 & 0.0550 & 0.0479 & \textbf{0.0623} \\
\midrule
w/o time embedding & 0.7197 & 0.0369 & 0.0017 & \textbf{0.0536} & 0.0481 & 0.0638 \\
w/o adaptivity & 0.7220 & 0.0548 & 0.0038 & 0.1028 & 0.0527 & 0.0638 \\
w/o sparse modulation & 0.2015 & 17.442 & \textbf{0.0007} & 0.650 & 0.0560 & 0.0626 \\
      \bottomrule
    \end{tabular}
  }

  \caption{\textbf{Motion inbetweening results} on the HML3D test set. Our CaKey design outperforms imputation sampling, LoRA, and CondMDI, highlighting the importance of context-aware modulation. Extensive ablations provided in Suppl.\S~{\color{red}{2}}.}
  \label{tab:mi_comparisons}
  \vspace{-10pt}
\end{table*}
\vspace{-10pt}
\subsection{Motion Inbetweening}\label{eval:mi}
\paragraph{\textbf{Quantitative results. }}We report quantitative comparisons between our first-stage model and other baselines, as summarized in Table~\ref{tab:mi_comparisons}. 
Simply applying imputation at inference yields suboptimal results, indicating that the generative prior of MDM cannot cover the sparsity of keyframes for inbetweening tasks, thus requiring further adaptation.
While using LoRA~\cite{hu2021lora} is effective, it still underperforms due to the lack of modules specifically designed for inbetweening. CondMDI~\cite{conmdi}, trained from scratch for inbetweening, also yields inferior results compared to ours. This result highlights the effectiveness of our CaKey design, whereas CondMDI merely concatenates keyframe masks with input motions, the CaKey layer leverages richer signals to modulate only the keyframe latents. 
Furthermore, we validate each design choice within the CaKey layer, with results showing that every component contributes critically to its overall effectiveness. 
Extensive ablations on CaKey can be found in the supplementary materials (Suppl.\S~{\color{red}{2}}).

\vspace{-5pt}
\paragraph{\textbf{Scene-awareness during inbetweening.}} 

\begin{wraptable}{r}{0.45\columnwidth}
\centering
\scriptsize
\begin{tabular}{clcc}
\toprule
\textbf{$s_k$} & \textbf{Stage.} &
\hphantom{000}\textbf{CFR$\downarrow$}\hphantom{000} & 
\hphantom{000}\textbf{MMP$\downarrow$}\hphantom{000} \\
\midrule
\multirow{2}{*}{20}
  & stage 1     & 0.021 & 0.011 \\
  & stage 2  & 0.022 \makebox[0pt][l]{\color{red}\scriptsize{($-5\%$)}} 
              & 0.010 \makebox[0pt][l]{\color{blue}\scriptsize{(+9\%)}} \\
\midrule
\multirow{2}{*}{40}
  & stage 1     & 0.030 & 0.016 \\
  & stage 2  & 0.030 \makebox[0pt][l]{\color{gray}\scriptsize{(0\%)}} 
              & 0.014 \makebox[0pt][l]{\color{blue}\scriptsize{(+13\%)}} \\
\midrule
\multirow{2}{*}{60}
  & stage1     & 0.037 & 0.020 \\
  & stage 2   & 0.033 \makebox[0pt][l]{\color{blue}\scriptsize{(+11\%)}} 
              & 0.015 \makebox[0pt][l]{\color{blue}\scriptsize{(+25\%)}} \\
\midrule
\multirow{2}{*}{80}
  & stage1     & 0.054 & 0.028 \\
  & stage 2  & 0.040 \makebox[0pt][l]{\color{blue}\scriptsize{(+26\%)}} 
              & 0.019 \makebox[0pt][l]{\color{blue}\scriptsize{(+32\%)}} \\
\bottomrule
\end{tabular}
\vspace{-3pt}
\caption{\textbf{Scene-awareness results on TRUMANS for inbetweening.}}
\vspace{-3pt}
\label{tab:samib}
\end{wraptable}
 While the keyframe stride $s_k$ is fixed at 20 in stage 1, we vary $s_k$ when training the SceneCo layer, and evaluate with $s_k$ used in stage 2 to examine improvements in scene-awareness on scene-aware inbetweening. As shown in Tab.~\ref{tab:samib}, using the same $s_k$ as stage 1 yields similar collision rates, since the model is already adapted specifically for motion inbetweening, leaving little room to improve. However, increasing $s_k$ encourages the scene-conditioning layer to more effectively exploit scene information, resulting in larger gains under sparser keyframe settings.

\vspace{-10pt}
\subsection{Analysis}\label{eval:analysis}
\vspace{-0pt}
\begin{figure}[t]
    \centering
    \begin{subfigure}{0.45\textwidth}
        \centering
        \includegraphics[width=\linewidth]{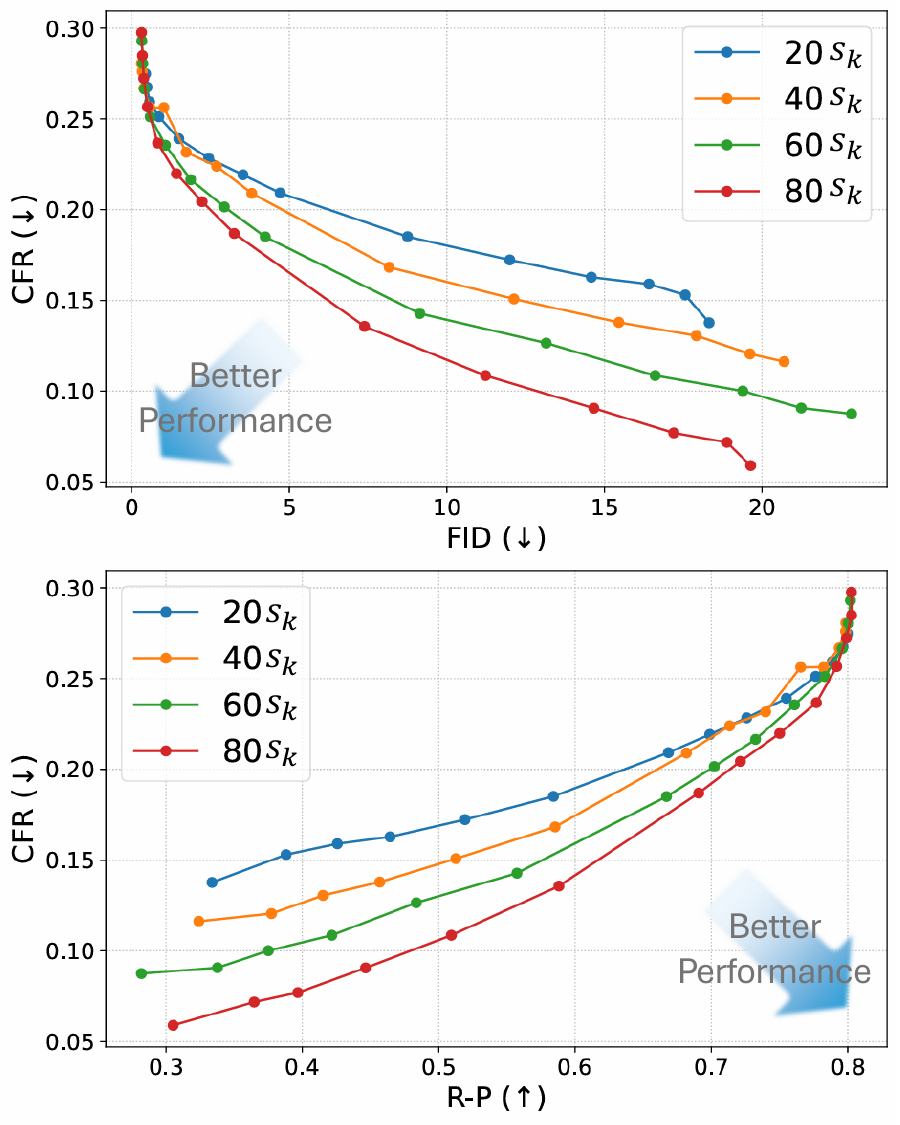}
        \caption{\textbf{Effect of Keyframe Strides ($s_k$).}}
        \label{fig:effect_of_KS}
    \end{subfigure}
    \begin{subfigure}{0.45\textwidth}
        \centering
        \includegraphics[width=\linewidth]{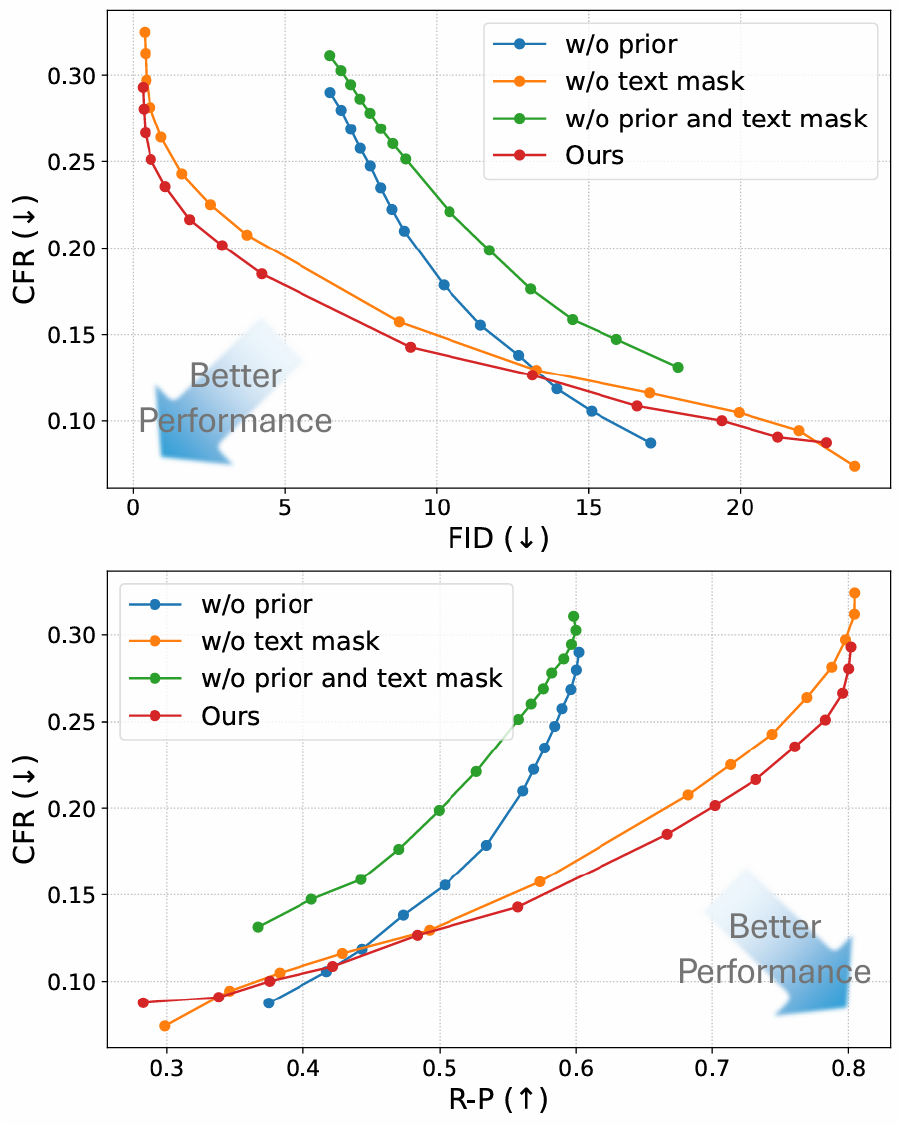}
        \caption{\textbf{Effect of Prior Preserving Designs.}}
        \label{fig:effect_of_prior_preservation}
    \end{subfigure}\hfill
\caption{\textbf{Ablation Studies.} Each dot represents a certain $w_s$ value during sampling, with higher values on the left and lower values on the right. For  
(a) : Sparser keyframes on stage 2 force the model to better exploit the scene, leading to an increase in scene-awareness.
(b) : The prior loss and the text mask for cross attention on stage 2 help the adaptation preserve the original T2M capabilities.
}
\end{figure}
\vspace{-5pt}
\paragraph{\textbf{Keyframe Stride at Stage 2.}} 
We ablate how varying $s_k$ in stage~2 impacts scene-aware text-to-motion performance. As shown in Fig.~\ref{fig:effect_of_KS}, sparser keyframes consistently improve performance, aligning with the results from scene-aware inbetweening (Tab.~\ref{tab:samib}). This indicates that the model leverages scene-awareness acquired in stage~2 and transfers it to text-conditioned motion generation.
\vspace{-10pt}
\paragraph{\textbf{Scene-conditioning Layer.}} 

\begin{figure}[t]
    \centering
    \includegraphics[width=\linewidth]{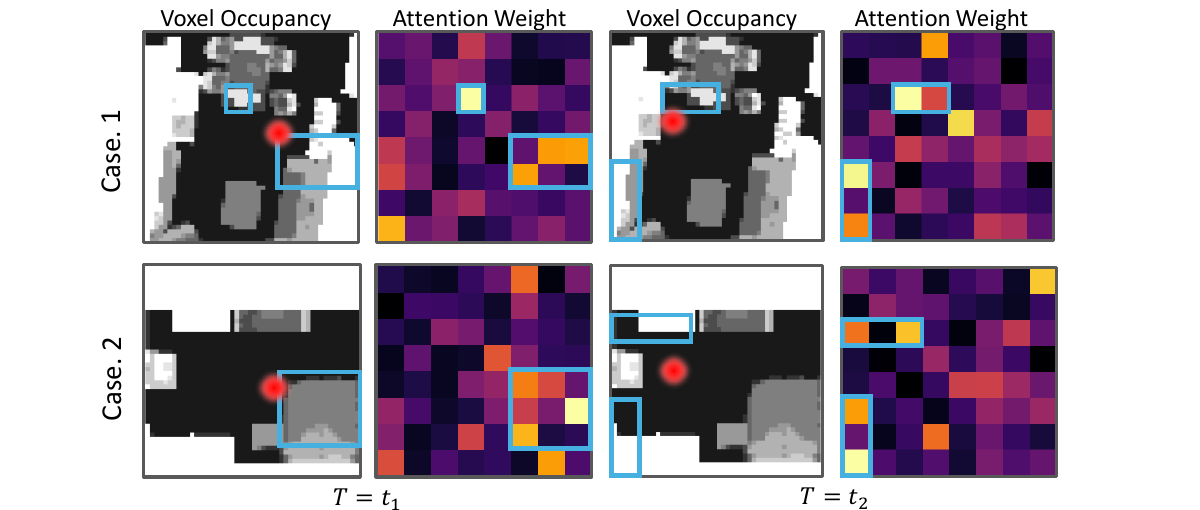}
    \vspace{-20pt}
    \caption{
\textbf{Visualization of cross-attention} weight maps between the motion latent at a given timestep and patch-wise scene embeddings.
The red point marks the human location. As highlighted by the blue boxes, the model predominantly attends to scene regions in the human’s immediate vicinity.}
    \label{fig:vis_scene_cross_attn}
    \vspace{-12pt} 
\end{figure}

\begin{wraptable}{r}{0.4\columnwidth}
\vspace{-5pt}
\centering
\scriptsize
\setlength{\tabcolsep}{4pt}
\renewcommand{\arraystretch}{1.05}
\begin{tabular}{lccc}
\toprule
\textbf{Emb. ($w_s$)} & \textbf{R-P$\uparrow$} & \textbf{CFR$\downarrow$} & \textbf{MMP$\downarrow$} \\
\midrule
Class (0) & 0.798 & 0.314 & 0.308 \\
Patch (0) & \cellcolor{gray!20}0.802 & \cellcolor{gray!20}0.293 & \cellcolor{gray!20}0.258 \\
\midrule
Class (0.1) & 0.799 & 0.306 & 0.299 \\
Patch (0.1) & \cellcolor{gray!20}0.800 & \cellcolor{gray!20}0.281 & \cellcolor{gray!20}0.242 \\
\midrule
Class (0.3) & \cellcolor{gray!20}0.796 & 0.290 & 0.281 \\
Patch (0.3) & 0.783 & \cellcolor{gray!20}0.251 & \cellcolor{gray!20}0.200 \\
\midrule
Class (0.4) & \cellcolor{gray!20}0.794 & 0.285 & 0.272 \\
Patch (0.4) & 0.761 & \cellcolor{gray!20}0.236 & \cellcolor{gray!20}0.178 \\
\bottomrule
\end{tabular}
\vspace{-5pt}
\caption{\textbf{Effect of Scene Rep.}}
\label{tab:effect_of_scene_rep}
\vspace{-3pt} 
\end{wraptable}
Using patch embeddings instead of class embeddings proves more effective for injecting scene-awareness, as shown in Tab.~\ref{tab:effect_of_scene_rep}. We further analyze attention weight maps of our scene-conditioning layer in Fig.~\ref{fig:vis_scene_cross_attn}. Occupied regions near the human receive high attention values, while empty regions nearby receive relatively low values. Moreover, attention weights vary dynamically along the human’s trajectory. These patterns suggest that motion latents interact with patch embeddings in a spatially adaptive manner through the cross-attention layers.
To assess robustness to different scene representations (e.g., meshes, TSDFs, and point clouds), we provide additional results in the supplementary materials (Suppl.\S~{\color{red}{4}}).
\vspace{-10pt}
\begin{wraptable}[5]{r}{0.40\columnwidth}
\centering
\scriptsize
\setlength{\tabcolsep}{2pt}
\renewcommand{\arraystretch}{1.05}
\begin{tabular}{lcc}
\toprule
\textbf{Strategy} & \textbf{FID$\downarrow$} & \textbf{R-P$\uparrow$} \\
\midrule
w/o inbetweening & 7.08 & 0.598 \\
Ours & \cellcolor{gray!20}0.497 & \cellcolor{gray!20}0.791 \\
\bottomrule
\end{tabular}
\vspace{-5pt}
\caption{\textbf{Effect of stage 1.}}
\vspace{-20pt}
\label{tab:effect_of_stage2}
\end{wraptable}
\paragraph{\textbf{Preserving Text-to-motion capabilities.} } 
 We analyze how the prior loss and text mask we introduce in stage 2 help preserve the original capabilities of MDM by evaluating on the scene-aware text-to-motion task.
 As shown in Fig.~\ref{fig:effect_of_prior_preservation}, incorporating the prior loss significantly improves T2M performance, and using the text mask during adaptation provides additional gains.
 To validate the necessity of our two-stage adaptation, we compare it with a single-stage variant that skips the motion inbetweening stage. As shown in Tab.~\ref{tab:effect_of_stage2}, without stage 1, the model loses its original text-to-motion capabilities, indicating that the inbetweening adaptation stage is crucial when learning scene-awareness from only scene-motion data.


\section{Discussion and Conclusion}
\vspace{-5pt}

We introduced \textbf{SceneAdapt}, a two-stage adaptation framework that injects scene-awareness into pretrained text-to-motion diffusion models. 
Our key idea is to use motion inbetweening as a bridge to leverage both text–motion and scene–motion datasets, avoiding the need for costly large-scale text–scene–motion collections.
In the first stage, the model is adapted for motion inbetweening through our Context-aware Keyframing (CaKey) layer, while in the second stage, scene-awareness is incorporated via scene-conditioning layers. Together, these adaptations enable the generation of motions that are both semantically rich and physically consistent with surrounding scenes. 
Extensive experiments confirm the effectiveness of each stage and validate the overall strength of the framework.

While effective, SceneAdapt has several limitations. First, although it generates diverse, scene-consistent motion, it does not capture dynamic scenes or object state changes (e.g., opening an oven while also generating the oven-door motion). Extending our framework to scene-aware human–object motion synthesis is a promising direction. Second, we do not model hand motion, which is critical for robotics and gaming; adapting whole-body models with articulated hands remains future work. Third, we focus on single-person motion; extending scene-aware generation to multi-human settings~\cite{hhoi} is an exciting avenue. Finally, our outputs are purely kinematic, and integrating physics-based control could yield more physically grounded, scene-aware behaviors~\cite{closd, Wu_2025_ICCV}.


%
%
\bibliographystyle{splncs04}
\bibliography{main}

\clearpage
\begin{center}
    {\Large \bfseries [Appendix]\\[0.5em] 
    SceneAdapt: Scene-aware Adaptation \\of Human Motion Diffusion \par}
\end{center}
\vspace{1em} 

\thispagestyle{plain}
\pagestyle{plain}

In this Appendix, we additionally provide:
\begin{itemize}
    \item \textbf{6. Evaluation set construction}
    \item \textbf{7. Detailed Ablations}
    \begin{itemize}
        \item A. Ablations on Cakey components
    \end{itemize}
    \item \textbf{8. Implementation Details}
    \begin{itemize}
        \item A. MDM Pretraining
        \item B. Inbetweening Stage
        \item C. Scene-Aware Inbetweening Stage
    \end{itemize}
    \item \textbf{9. Results}
    \begin{itemize}
        \item A. Distribution of Penetration-Depth
        \item B. Different Scene Representations
    \end{itemize}
    \item \textbf{10. Positioning}
    \item \textbf{11. Use of Large Language Models}
\end{itemize}

\section{Evaluation set construction}\label{appendix_a}
We construct our evaluation set of text--scene--motion pairs as follows. First, we extract the 3D pelvis joint coordinates and their corresponding signed distance function (SDF) values from each frame in the TRUMANS dataset, utilizing SDF fields precomputed from the provided scene meshes. We discard frames where the root height indicates a sitting or lying posture. 

\begin{figure}[h]
    \centering
    \includegraphics[width=0.96\textwidth]{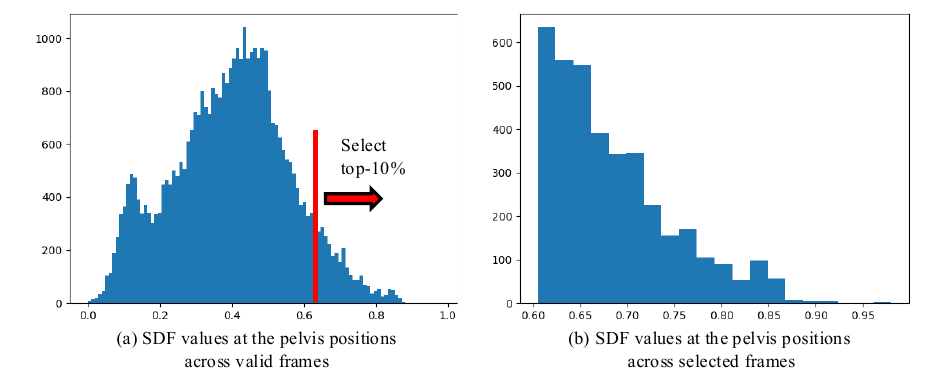}
\caption{\textbf{Distribution of SDF values at pelvis positions in the evaluation set.} The x-axis denotes the SDF value, and the y-axis denotes the number of frames. (a) shows the distribution over valid frames after filtering out sitting and lying postures. (b) shows the histogram after selecting the top \(10\%\) of frames by SDF value. The selected pelvis positions have SDF values ranging from 
\(0.6\,\mathrm{m}\) to \(1.0\,\mathrm{m}\), providing physically plausible clearance for realistic human motion.}
    \label{fig:supp_eval_set_sdf_hist}
\end{figure}

From the remaining data, we filter out frames with low SDF values, as these represent starting positions too close to surrounding objects and can lead to physically implausible motion synthesis. For example, it is unnatural to generate motion when ``a person runs forward'' is given as the text condition, but the starting point is already immediately in front of a wall. Specifically, we retain only the top 10\% of frames with the highest SDF values to guarantee ample clearance between the starting point and the surrounding scene. As shown in Fig~\ref{fig:supp_eval_set_sdf_hist}, the SDF values at these selected positions range from \(0.6\,\mathrm{m}\) to \(1.0\,\mathrm{m}\), ensuring sufficient physical space for realistic human motion.

To provide motion and text annotations, we randomly sample sequences from HumanML3D. We explicitly exclude climbing or stair-related actions, as these require specific architectural features absent from the TRUMANS scenes. Below are examples of the motion descriptions excluded from our dataset:

\begin{itemize}
    \item ``a person starts at the top of stairs and at a regular pace walks down stairs.''
    \item ``a person walks down some stairs.''
    \item ``a person stomps up some stairs.''
    \item ``a person walks up a staircase, then stops.''
    \item ``the person turns to their left and begins walking up what seem to be 3 stair steps.''
    \item ``a person slowly walks forward and then up a flight of stairs.''
    \item ``a person quickly climbs a flight of stairs.''
\end{itemize}

Following this procedure, we obtain a final evaluation set of 3K text--scene--motion pairs.

\section{Detailed Ablations}\label{appendix_b}
\begin{table*}[h]
\centering
\setlength{\tabcolsep}{4pt}
\renewcommand{\arraystretch}{1.2}
\resizebox{\textwidth}{!}{%
\begin{tabular}{ccccccc}
\toprule
\textbf{Sparse Mod.} &
\textbf{Adaptive} &
\textbf{Time emb.} &
\textbf{Modulator} &
\textbf{FID}$\downarrow$ & 
\textbf{MJPE (Key)}$\downarrow$ & 
\textbf{MJPE(All)}$\downarrow$ \\
\midrule

\cmark & \cmark & \cmark & MLP & 0.0356 & 0.0018 & 0.055 \\
 & \cmark & \cmark & MLP & 17.442 & 0.0007 & 0.650 \\
\cmark & \cmark & & MLP & 0.0369 & 0.0017 & 0.0536 \\
\cmark & & \cmark & MLP &  0.0548 & 0.0038 & 0.1028 \\
\cmark & \cmark & \cmark & Linear & 0.0485 & 0.0027 & 0.0764 \\
\cmark & & \cmark & Linear & 0.0924 & 0.0051 & 0.1308 \\
\cmark & \cmark & & Linear & 0.0485 & 0.0027 & 0.0764 \\
\cmark & & & Linear & 0.0849 & 0.0044 & 0.1173 \\
\bottomrule
\end{tabular}}
\vspace{5pt}
\caption{\textbf{Ablation study on motion inbetweening designs.} 
Sparse Mod. indicates whether sparse modulation is used. 
Adaptive denotes whether the source latent is provided as input to the modulator. 
Time emb. specifies whether time embedding is provided as input to the modulator. 
Modulator describes how $f_{\theta}$ and $h_{\phi}$ are modeled.}
\vspace{-10pt}
\label{tab:mi_ablation}
\end{table*}
\textbf{Ablations on Cakey components. }As reported at Table~\ref{tab:mi_ablation}, We ablate key components of CaKey layer introduced in \ref{sec:inbetween}.
One crucial element is the sparse modulation, which focuses on keyframe poses while preserving the non-keyframe latents.
Replacing it with global modulation (second row) results in a significant performance drop, validating its effectiveness.
As shown in the third and fourth rows, leveraging contextual signals such as source latent motion or timestep embeddings is also critical.
Finally, the network design of the modulator is important for fully utilizing these contexts, as models with MLPs consistently outperform those with linear layers.

\section{Implementation Details}\label{appendix_c}

\textbf{MDM Pretraining.} As shown in \cite{conmdi}, using motion representations with global root information can lead to severe foot skating results, which can be alleviated by adopting a U-Net architecture~\cite{gmd} instead of the transformer architecture originally used in MDM. In our experiments, we found that introducing additional global position and velocity losses significantly improves motion naturalness, achieving the same performance to the original motion representation used in \cite{humanml3d}. We therefore pretrain MDM using the following losses:
\begin{align}
\mathcal{L}_{\text{joints}} &=
\mathbb{E}_{x_0 \sim q(x_0 \mid \mathcal{T}),\, t \sim [1, T]}
\Big[ \big\| \mathrm{FK}(x_0) - \mathrm{FK}(\mathcal{D}_\theta(x_t, t, \mathcal{T})) \big\|_2^2 \Big], \\
\mathcal{L}_{\text{vel}} &=
\mathbb{E}_{x_0 \sim q(x_0 \mid \mathcal{T}),\, t \sim [1, T]}
\Big[ \big\| \mathrm{diff}(\mathrm{FK}(x_0)) - \mathrm{diff}(\mathrm{FK}(\mathcal{D}_\theta(x_t, t, \mathcal{T}))) \big\|_2^2 \Big].
\end{align}

\noindent
where $\mathrm{FK}$ denotes forward kinematics, and $\mathrm{diff}$ refers to the temporal difference of the joint positions. The total loss is given by:
\begin{equation}\label{equation:final_loss}
\mathcal{L} =
\mathcal{L}_{\text{t2m}} +
\lambda_{\text{joints}} \mathcal{L}_{\text{joints}} +
\lambda_{\text{vel}} \mathcal{L}_{\text{vel}},
\end{equation}
where $\lambda_{\text{joints}} = 1$ and $\lambda_{\text{vel}} = 100$.

\textbf{Inbetweening Stage.} For the CaKey layers, we use a single-layer MLP with SiLU activations, initialized such that the modulation does not affect the latents at the start of adaptation. Each layer modulates the latents after the self-attention block within each transformer block of MDM. We train for 200k steps with a learning rate of $1\times10^{-4}$ using the AdamW optimizer. The same loss functions used in MDM pretraining are applied.

\textbf{Scene-Aware Inbetweening Stage.} For the voxel feature extractor, we employ a 512-dimensional ViT with 4 layers and 4 attention heads, using a patch size of 6 to produce 64 patches in total. Scene-conditioning layers are added to all transformer layers of MDM, where cross-attention is applied immediately after the CaKey layers. To stabilize adaptation, we apply layer normalization to both the key–value pairs and the query, and use gradient clipping. Training proceeds for 200k steps with the same loss weights as the text-to-motion stage.
\begin{figure}[t]
    \vspace{-5pt} 
    \centering
    \includegraphics[width=0.96\textwidth]{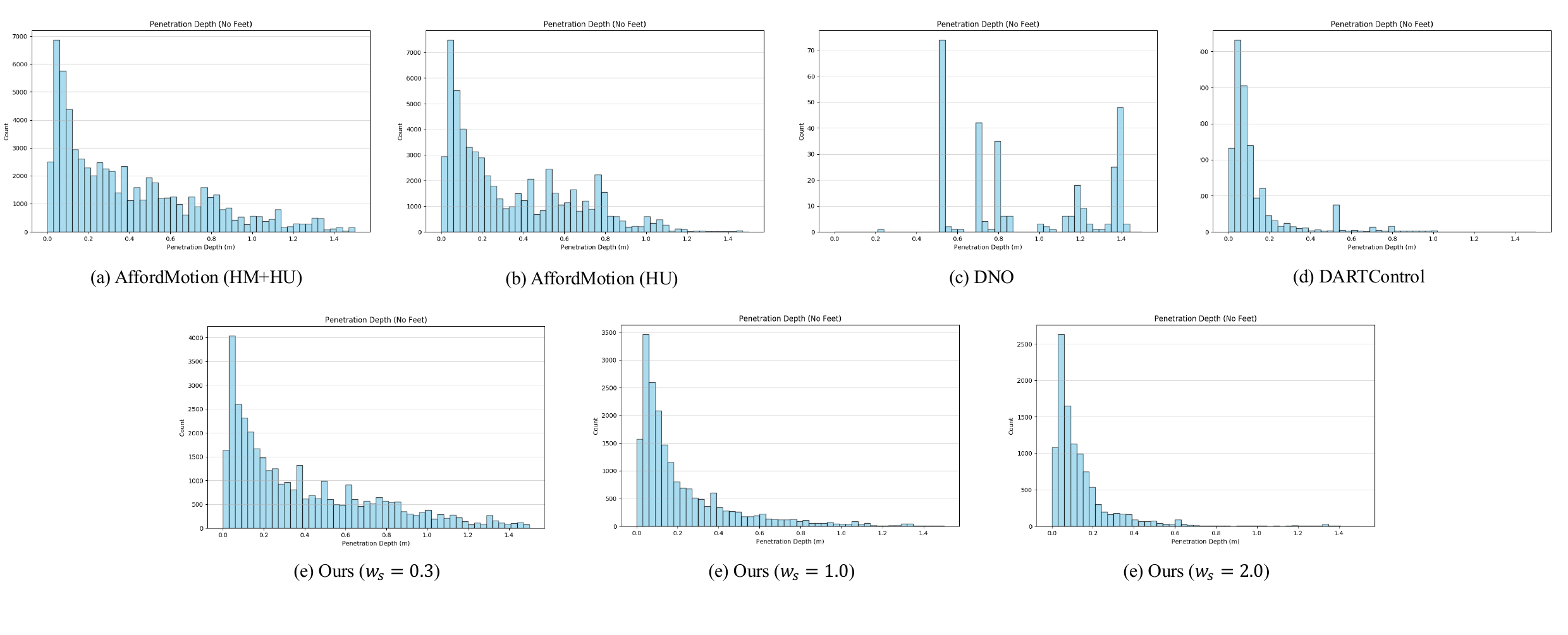}
    \vspace{-15pt}
\caption{\textbf{Distribution of Penetration-Depth.} This figure shows the distribution of the per-frame maximum penetration depth over colliding frames (unit: m). Note that the total count differs across histograms because frames without any penetration are excluded from this plot. Please zoom in to see the details.}
    \label{fig:supp_ppr_dist}
\end{figure}
\begin{table}[t]
\centering
\small
\setlength{\tabcolsep}{4pt}
\begin{tabular}{lccc}
\toprule
Method & ZR (\%) $\uparrow$ & P50 $\downarrow$ & P90 $\downarrow$ \\
\midrule
DNO                & 99.3 & 0.80 & 1.39 \\
DARTControl        & 98.2 & 0.08 & 0.31 \\
AffordMotion (HU)  & 43.1 & 0.23 & 0.78 \\
AffordMotion (all) & 37.2 & 0.29 & 1.87 \\
\textbf{Ours $w_s=0.3$}      & 64.0 & 0.28 & 0.99 \\
\textbf{Ours $w_s=1.0$}      & 80.6 & 0.13 & 0.57 \\
\textbf{Ours $w_s=2.0$}      & 89.7 & 0.08 & 0.32 \\
\bottomrule
\end{tabular}

\vspace{5pt} 

\caption{\textbf{Comparison of penetration statistics across methods.} ZR denotes the fraction (\%) of penetration-free frames among all evaluation frames (higher is better), while P50 and P90 are the 50th and 90th percentiles of the per-frame maximum penetration depth computed over colliding frames only (lower is better).}
\label{tab:ppr_dist_percentile}
\vspace{-10pt}
\end{table}
\section{Results}
\begin{table}[t]
\centering
\small
\begin{tabular}{lccccc}
\toprule
\textbf{Method / $w_s$} & \textbf{R-P (top 3) ↑} & \textbf{FID ↓} & \textbf{CFR ↓} & \textbf{MMP ↓} & \textbf{JCR ↓} \\
\midrule
point cloud ($w_s$ = 0)     & 0.803 & 0.395 & 0.297 & 0.447 & 0.306 \\
point cloud ($w_s$ = 0.3)   & 0.800 & 0.394 & 0.290 & 0.433 & 0.300 \\
point cloud ($w_s$ = 0.5)   & 0.795 & 0.400 & 0.289 & 0.426 & 0.296 \\
point cloud ($w_s$ = 2.0)   & 0.667 & 3.93  & 0.275 & 0.397 & 0.280 \\
\midrule
TSDF ($w_s$ = 0)            & 0.801 & 0.347 & 0.292 & 0.425 & 0.295 \\
TSDF ($w_s$ = 0.3)          & 0.786 & 0.545 & 0.259 & 0.347 & 0.257 \\
TSDF ($w_s$ = 0.5)          & 0.748 & 1.44  & 0.231 & 0.283 & 0.217 \\
TSDF ($w_s$ = 2.0)          & 0.359 & 20.97 & 0.095 & 0.078 & 0.063 \\
\midrule
mesh ($w_s$ = 0)            & 0.801 & 0.471 & 0.311 & 0.503 & 0.333 \\
mesh ($w_s$ = 0.3)          & 0.800 & 0.489 & 0.305 & 0.486 & 0.327 \\
mesh ($w_s$ = 0.5)          & 0.798 & 0.508 & 0.302 & 0.475 & 0.322 \\
mesh ($w_s$ = 2.0)          & 0.747 & 1.47  & 0.267 & 0.381 & 0.274 \\
\midrule
voxel ($w_s$ = 0)           & 0.803 & 0.312 & 0.298 & 0.273 & 0.299 \\
voxel ($w_s$ = 0.3)         & 0.792 & 0.497 & 0.256 & 0.208 & 0.246 \\
voxel ($w_s$ = 0.5)         & 0.750 & 1.42  & 0.220 & 0.160 & 0.199 \\
voxel ($w_s$ = 2.0)         & 0.365 & 18.88 & 0.072 & 0.035 & 0.045 \\
\bottomrule
\end{tabular}
\vspace{5pt}
\caption{\textbf{Effect of different scene representations.}}
\end{table}\label{tab:diverse_representation}
\textbf{Distribution of Penetration-Depth.} To provide further detailed performance about scene-awareness, we visualize the comparisons of penetration-depth distribution as shown in Fig~\ref{fig:supp_ppr_dist}.
Compared to AffordMotion, our method yields a penetration-depth histogram that is more concentrated near zero, and the bin counts indicate that our motions exhibit fewer penetrations overall. 
Optimization-based baselines (DNO, DARTControl) also show low penetration, but closer inspection suggests that this result mainly stems from their limited motion or poor adherence to the input text.
These tendencies can also be seen quantitatively in Table~\ref{tab:ppr_dist_percentile}.

\textbf{Different Scene Representations.}
We further investigate SceneAdapt’s robustness to different forms of scene encoding. In addition to our default voxel-based representation, we experiment with three alternative scene encodings: point clouds, TSDF volumes, and meshes. For point clouds, we adopt a Point Transformer encoder; for meshes, a Mesh Transformer encoder; and for TSDF volumes, we reuse the same voxel-ViT encoder employed in our default model. To assess the effect of classifier-free guidance for scene conditioning, we evaluate each representation using multiple scene CFG weights. The full quantitative comparison is reported in Table~\ref{tab:diverse_representation}. Across all evaluations, we observe that voxel-based scene representations yield the strongest and most stable performance. Their dense and spatially regular structure provides a rich geometric signal that aligns well with SceneAdapt’s conditioning architecture, resulting in both high semantic fidelity and strong geometric consistency. TSDF volumes perform competitively, offering similar advantages with slightly smoother geometric fields. In contrast, unstructured or sparse representations such as point clouds and meshes are less effective: although they still enable scene-aware behavior, their irregular sampling and lower spatial density provide weaker geometric cues, limiting their ability to enforce scene constraints.

\begin{table}[t]
\centering
\resizebox{\columnwidth}{!}{
\begin{tabular}{lcccc}
\toprule
\textbf{Dataset} & 
\textbf{Motion Semantic} & 
\textbf{Duration (min)} & 
\textbf{Scene} & 
\textbf{Open Source} \\
\midrule
HumanML3D      
    & Diverse 
    & 1715 
    & \xmark 
    & \cmark \\

HUMANISE       
    & Limited (4 actions) 
    & 600 (purely 51) 
    & \cmark\ (Synthetic) 
    & \cmark \\

Trumans        
    & Limited (10 actions) 
    & 900 
    & \cmark 
    & \cmark \\

LaserHuman     
    & Moderate 
    & 180 
    & \cmark 
    & \xmark \\

SAMP           
    & Limited (5 actions) 
    & 103 
    & \xmark\ (Object-only) 
    & \cmark \\
\bottomrule
\end{tabular}
}
\vspace{5pt}
\caption{Comparison of datasets used for scene-aware motion generation.}\label{tab:dataset}
\vspace{-10pt}
\end{table}

\begin{table}[t]
\centering
\resizebox{\columnwidth}{!}{
\begin{tabular}{lccccc}
\toprule
\textbf{Method} & 
\begin{tabular}{c}\textbf{Rich Motion}\\[-2pt]\textbf{Semantics}\end{tabular} &
\begin{tabular}{c}\textbf{Scene Geometry}\\[-2pt]\textbf{Awareness}\end{tabular} &
\begin{tabular}{c}\textbf{No Triplet}\\[-2pt]\textbf{Needed}\end{tabular} &
\begin{tabular}{c}\textbf{Learned From}\\[-2pt]\textbf{GT Interactions}\end{tabular} &
\begin{tabular}{c}\textbf{Open}\\[-2pt]\textbf{Source}\end{tabular} \\
\midrule
Humanise CVAE~\cite{humanise}       
    & \xmark & \cmark & \xmark & \xmark & \cmark \\

Afford Motion~\cite{affordmotion}   
    & \xmark\ (degrades due to HUMANISE) & \cmark & \xmark & \xmark & \cmark \\

Cen et al.~\cite{cen2024text_scene_motion}  
    & \xmark & \cmark & \xmark & \xmark & \cmark \\

TeSMo~\cite{yi2024tesmo}            
    & \xmark & \cmark & \xmark & \cmark\ (interaction) / \xmark\ (locomotion) & \cmark \\

LaserHuman~\cite{cong2024laserhuman} 
    & \xmark & \cmark & \xmark & \cmark & \xmark \\

\textbf{Ours (SceneAdapt)}          
    & \textbf{\cmark} & \textbf{\cmark} & \textbf{\cmark} & \textbf{\cmark} & \textbf{\cmark} \\
\bottomrule
\end{tabular}
}
\vspace{5pt}
\caption{Comparison of scene-aware motion generation methods.}\label{tab:positioning}
\end{table}

\section{Positioning}
In this section, we provide a detailed comparison between SceneAdapt and related works. A key motivation behind SceneAdapt is the absence of large-scale text--scene--motion datasets that exhibit rich semantic diversity. As demonstrated in Table~\ref{tab:dataset}, existing datasets fail to simultaneously provide semantic diversity and scene awareness. For instance, while HumanML3D (HML3D) features diverse semantics, it lacks scene context; conversely, datasets containing scene information typically offer limited text annotations. Although LaserHuman provides moderate diversity, it comprises only 180 minutes of motion data and remains closed-source, with only a small set of demonstration samples publicly available. This limitation in the data landscape fundamentally shapes our problem formulation and explains why prior methods struggle to synthesize motions that are both semantically expressive and physically consistent with the scene. 
As summarized in Table~\ref{tab:positioning}, most existing methods lack diverse motion semantics because they strictly rely on paired text--scene--motion triplet data. Furthermore, many of these works resort to learning from pseudo-annotated scene--motion datasets, such as HUMANISE. In contrast, SceneAdapt is uniquely capable of generating semantically diverse motions that strictly adhere to scene geometry. This is achieved through our novel formulation, which effectively circumvents the need for triplet training data.

\section{Use of Large Language Models}
We only utilized Large Language Models to polish our written draft.

\end{document}